%% file: colm2026_conference.tex
\documentclass{article} 
\usepackage[preprint]{colm2026_conference}

\usepackage{microtype}
\usepackage{hyperref}
\usepackage{url}
\usepackage{booktabs}
\usepackage{graphicx}
\usepackage{booktabs}
\usepackage{multirow}
\usepackage{amsfonts}
\usepackage{subcaption}
\usepackage{tcolorbox}
\definecolor{mydarkpink}{HTML}{c2255c}
\definecolor{mypink}{HTML}{e64980}
\usepackage{amsmath} 
\usepackage{makecell}
\usepackage{adjustbox}

\usepackage{microtype}
\usepackage{xspace}
\usepackage{subcaption}
\usepackage{booktabs} %
\usepackage{arydshln}
\usepackage{algorithm}
\usepackage{mathrsfs}
\usepackage{setspace}
\usepackage[noend]{algpseudocode}
\usepackage{amsmath}
\usepackage{amssymb}
\usepackage{mathtools}
\usepackage{amsthm}
\usepackage{booktabs} 
\usepackage[inkscapelatex=false]{svg}
\usepackage{thmtools} %
\usepackage{caption} %
\usepackage{multirow} %
\usepackage{float}
\newcommand{\method}{\texttt{MR-Search}\xspace}

\usepackage{cuted}   

\newcommand{\think}[1]{\textcolor{blue}{\texttt{<think>}} #1 \textcolor{blue}{\texttt{</think>}}}
\newcommand{\search}[1]{\textcolor{cyan}{\texttt{<search>}} #1 \textcolor{cyan}{\texttt{</search>}}}
\newcommand{\info}[1]{\textcolor{brown}{\texttt{<information>}} #1 \textcolor{brown}{\texttt{</information>}}}
\newcommand{\answer}[1]{\textcolor{purple}{\textbf{\texttt{<answer>}}} {\textbf{#1}} \textcolor{purple}{\textbf{\texttt{</answer>}}}}
\newcommand{\prompt}[1]{\textcolor{darkgray}{\textbf{#1}}}

\usepackage{lineno}
\usepackage{wrapfig}
\definecolor{darkblue}{rgb}{0, 0, 0.5}
\hypersetup{colorlinks=true, citecolor=darkblue, linkcolor=darkblue, urlcolor=darkblue}

\title{Meta-Reinforcement Learning with \\ Self-Reflection for Agentic Search}

\author{
    Teng Xiao$^{1,2}$\thanks{Equal contribution. Correspondence to \texttt{tengx@allenai.org}.$^\dagger$Equal senior authors.} \; 
    Yige Yuan$^{2*}$\; 
    Hamish Ivison$^{1,2}$\; 
    Huaisheng Zhu$^{3}$\;
    Faeze Brahman$^{1}$  \\
    \textbf{
    Nathan Lambert$^{1}$\;
    Pradeep Dasigi$^{1,\dagger}$\; 
    Noah A. Smith$^{1,2,\dagger}$\; 
    Hannaneh Hajishirzi$^{1,2,\dagger}$} \\ 
    \;$^1$Allen Institute for AI\;
    $^2$University of Washington\;
    $^3$Independent \
}

\begin{document}

\ifcolmsubmission
\linenumbers
\fi

\maketitle

\begin{abstract}
This paper introduces \method, an in-context meta reinforcement learning (RL) formulation for agentic search with self-reflection. Instead of optimizing a policy within a single independent episode with sparse rewards, \method trains a policy that conditions on past episodes and adapts its search strategy across episodes.  \method \emph{learns to learn} a search strategy with self-reflection, allowing search agents to improve in-context exploration at test-time. Specifically, \method performs cross-episode exploration by generating explicit self-reflections after each episode and leveraging them as additional context to guide subsequent attempts, thereby promoting more effective exploration during test-time. We further introduce a multi-turn RL algorithm that estimates a dense relative advantage at the turn level, enabling fine-grained credit assignment on each episode. Empirical results across various benchmarks demonstrate the advantages of \method over baselines based RL, showing strong generalization and relative improvements of \textbf{9.2}\% to \textbf{19.3}\% across eight benchmarks. Our code and data are available at \url{https://github.com/tengxiao1/MR-Search}.
\end{abstract}

\input{sections/introduction}

\input{sections/related}
\input{sections/method}

\input{sections/experiments}

\input{sections/discussion}
\input{sections/conclusion}

\section*{Acknowledgment}
This material is based upon work supported by the National Science Foundation under Award No. 2413244.
The views expressed are those of the author and do not reflect the official policy or position of the Department of Defense or the U.S. Government.

\section*{Limitations}
Despite its effectiveness, \method has several limitations. \textit{First}, we do not evaluate our method on long-form benchmarks, where responses are substantially longer. Verification in such settings is inherently challenging, and how to reliably assess progress and final correctness for long-form generation remains an open research question. \textit{Second}, our current study focuses on agentic search with a fixed Wikipedia search tool. Extending \method to environments involving multiple heterogeneous tools, such as combined web search and web browsing. We leave the investigation of these directions to future work. It would be
also particularly interesting to scale \method to large agentic RL training runs and further study the scaling properties of Meta-RL with frontier base models.

\section*{Ethics Statement}
This paper advances reinforcement learning for agentic search by improving training efficiency and credit assignment under sparse rewards. By enabling more effective training of language-model-based agents, this work may benefit practical applications that require efficient and reliable reasoning. This research shares the societal implications of machine learning systems more broadly and does not introduce additional ethical concerns beyond those commonly associated with large language models and reinforcement learning.

\bibliography{colm2026_conference}
\bibliographystyle{colm2026_conference}

\newpage
\appendix
\section{Appendix}
\input{sections/appendix}

\end{document}

%% file: sections/introduction.tex
\section{Introduction}
Language models with advanced reasoning capabilities have driven substantial progress toward more autonomous and multi-step decision-making behaviors in complex tasks~\citep{guo2025deepseek,jaech2024openai}. Examples include agentic search such as deep research~\citep{du2025deepresearch,shao2025dr} and other information seeking~\citep{mialon2023gaia,jin2025search}, where LMs use search tools and engage in dynamic, multi-turn interactions. Reinforcement learning (RL) with the ReAct paradigm \citep{yao2022react} has emerged as a primary framework for training search agents~\citep{jin2025search,zheng2025deepresearcher}, replacing the traditional reliance on supervised data collection. Specifically, these methods mainly focus on the correctness of the final answer and only receive sparse rewards at the end of each trajectory, without providing precise credit assignment for intermediate steps.  Due to the sparse nature of outcome rewards, the agent often struggles to learn more complex processes and is susceptible to issues such as inefficient exploration at the early stage, local optima, and inefficient search dynamics~\citep{zhang2025agent,feng2025group}. These challenges become more pronounced in agentic search tasks, where  multi-turn interactions with tools could amplify small errors and obscure credit assignment~\citep{feng2025group}.

To address the key challenge of sparse rewards, several works have explored using \emph{process} reward models~\citep{luo2024improve,wang2023math} or LM judges~\citep{zheng2024processbench,deng2025atom}. However, these approaches rely on external annotations, which are both costly and difficult to reuse when task requirements change. Moreover, model-based rewards inevitably lead to reward hacking and bias~\citep{wang2025towards} and incur additional computational overhead in the RL training~\citep{yuan2024free}.

\begin{figure}[!h]
\centering
\includegraphics[width=0.95\linewidth]{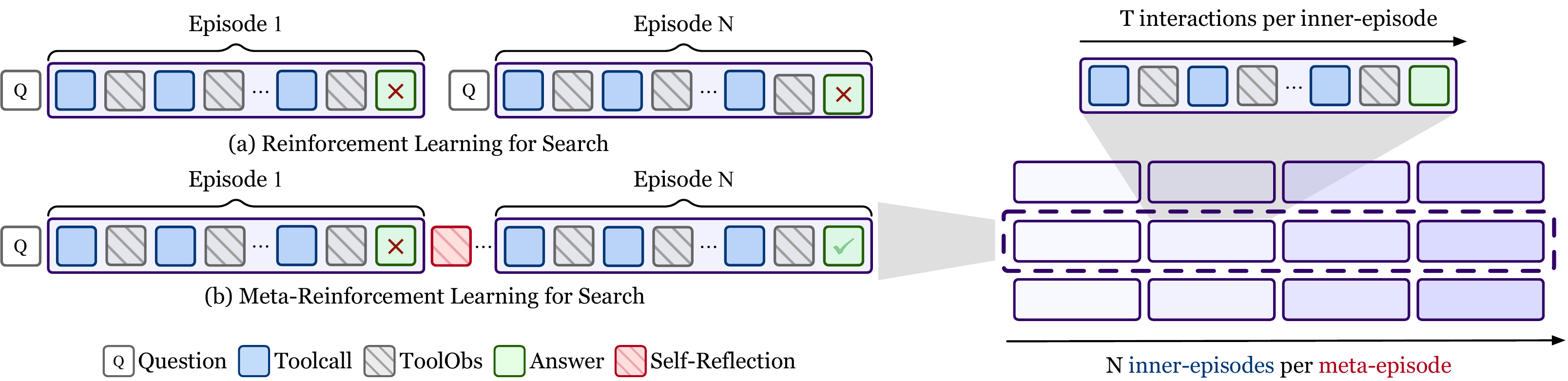}
\caption{RL-based agents (a) condition solely on the current episode, and episodes are \emph{independent}, whereas meta-RL-based agents (b) leverage context accumulated across episodes. \method performs sequential self-reflection over past episodes to guide exploration in subsequent episodes. In \method, we have inner-episodes, each consisting of a maximum of $T$ interactions steps with an answer. A sequence of $N$ episodes forms a meta-episode.}
\label{fig:paradigm} 
\vspace{-1.5em}
\end{figure}

In this paper, we introduce \method, a simple yet effective meta-RL approach that enables search agents to improve in-context exploration at test time. Our work is most closely related to in-context meta-reinforcement learning  methods~\citep{duan2016rl,stadie2018some,laskincontext}, which leverage in-context histories from a few initial exploration episodes to guide subsequent exploitation episodes. Unlike traditional meta-RL approaches~\citep{duan2016rl,stadie2018some, laskincontext} in robotics and games, we focus on open-domain agentic search tasks with tool interactions and self-reflection, without any reward feedback from the environment during inference.

As illustrated in Figure~\ref{fig:paradigm}, \method formulates agentic search as an iterative, self-reflection driven process instead of performing exploration through multiple independent episodes that operate in isolation in RL-base agents. In \method, each complete interaction trajectory with an answer is an episode and followed by an explicit reflection step. This design enables sequential self-reflection and cross-episode knowledge consolidation in a multi-turn setting, transforming exploration from a set of disconnected attempts into a progressively informed search process. Thus, the agent learns to balance exploration and exploitation end-to-end by updating its search strategy according to final task performance. Our method can be seen as an instance of meta-learning where we meta-learn how to generate effective self-reflection. 

To optimize the policy with multi-turn reflection, we use a multi-turn RL algorithm that estimates unbiased grouped relative advantages~\citep{ahmadian2024back} at the turn level to assign localized credit to self-reflection turns. As a result, \method remains critic-free and eliminates the need for auxiliary value models compared to PPO~\citep{schulman2017proximal}.

Our main contributions are:
(i) We advocate for and formalize in-context meta-reinforcement learning as a practical and scalable bridge between meta-learning and reinforcement learning for agentic search, where ground-truth rewards are absent at inference time.
(ii) We propose \method, an effective multi-turn agentic search framework that performs cross-episode exploration by generating an explicit self-reflection after each interaction episode.
(iii) Empirically, we validate the effectiveness of \method across multiple multi-hop QA benchmarks, showing that it significantly outperforms prior methods. Specifically, \method achieves an average relative improvement of \textbf{9.2}\% to \textbf{19.3}\% over strong baselines.

%% file: sections/related.tex
\begin{section}{Related Work}\label{sec:related}

\textbf{RL for Agentic Search.}  RL has emerged as a promising training paradigm for developing adaptive and autonomous search agents~\citep{jin2025search,wu2025webdancer,li2025websailor}. Specifically, agentic search with RL trains LLMs as decision-making agents that interact with a search environment through reasoning, receive feedback, and iteratively refine their strategies to maximize task rewards. For instance, Search-R1 and ReSearch~\citep{jin2025search,chen2025learning} propose training LLM-based agents end-to-end using RL algorithms such as PPO~\citep{schulman2017proximal} or GRPO~\citep{shao2024deepseekmath} under the ReAct paradigm~\citep{yao2022react}. Despite significant progress, these methods rely solely on sparse outcome rewards, without providing precise credit assignment for effective exploration~\cite{feng2025group}. Recently, several works~\citep{deng2025atom,wang2025stepsearch} have attempted to design  process rewards for agentic search. However, these approaches require annotations at every intermediate step or rely on external evaluators, both of which are expensive to obtain. In contrast, our \method aligns with in-context meta-reinforcement learning, enabling progressively targeted exploration driven by explicit cross-episode reflection.

\textbf{Meta-Reinforcement Learning.} Our work is related to meta-RL methods that leverage in-context histories to guide exploration, with the goal of maximizing rewards in subsequent exploitation episodes in robotics and game domains~\citep{melo2022transformers,laskincontext}. \citet{duan2016rl} and \citet{wang2017learning} concurrently proposed the RL$^2$ framework, which formulates meta-reinforcement learning by feeding in-context episodes into a recurrent neural network (RNN) whose hidden state serves as a memory mechanism.  \citet{qu2025optimizing} formulate the optimization of test-time compute for LLMs as a meta-reinforcement learning problem. A concurrent work~\citep{jiang2025meta} proposes meta-RL to encourage exploration in LLM agents with ground-truth state feedback. In contrast, we focus on LLM open-domain agentic  task  and do not access to any environment feedback during inference.

\textbf{LLMs with Self-Reflection.} 
Previous studies investigate prompting-based strategies that enable LLMs to iteratively and sequentially refine their own generations via intrinsic feedback, effectively scaling test-time computation. These approaches include self-correction~\citep{huang2023large}, self-reflection~\citep{shinn2023reflexion}, and self-refine~\citep{madaan2023self}. Beyond prompting alone, several subsequent works~\citep{wan2025srpo,qu2024recursive,kumar2024training,xiong2025self, yuksekgonul2026learning} employ finetuning to train models for intrinsic self-correction. A recent concurrent work~\citep{shi2026experiential} also proposes experiential reinforcement learning to incorporate textual reflection feedback. In contrast, our work offers a novel meta-RL perspective on agentic search with tool interaction to catalyze continuous self-reflection, enabling the model to more effectively explore better answers.

\textbf{Test-time Scaling.}
The paradigm of scaling test-time compute has emerged as a critical avenue for enhancing reasoning capabilities~\citep{guo2025deepseek}. Prior work explores two main dimensions: parallel sampling~\citep{wang2022self} and sequential refinement~\citep{madaan2023self,snell2024scaling}. Parallel sampling generates multiple answers independently, whereas sequential refinement generates answers sequentially, with each attempt conditioned on previous ones. Our work is most closely related to the sequential refinement paradigm. However, unlike these methods, which operate purely at inference time, this work proposes cross-episode meta-learning to enable continuous self-reflection in context, showing that meta-RL induces effective in-context exploration for agentic search.

\end{section}

%% file: sections/method.tex
\section{MetaSearch: Meta-Reinforcement Learning for Agentic Search}\label{sec:method}

\begin{figure}[!t]
\centering
\includegraphics[width=0.9\linewidth]{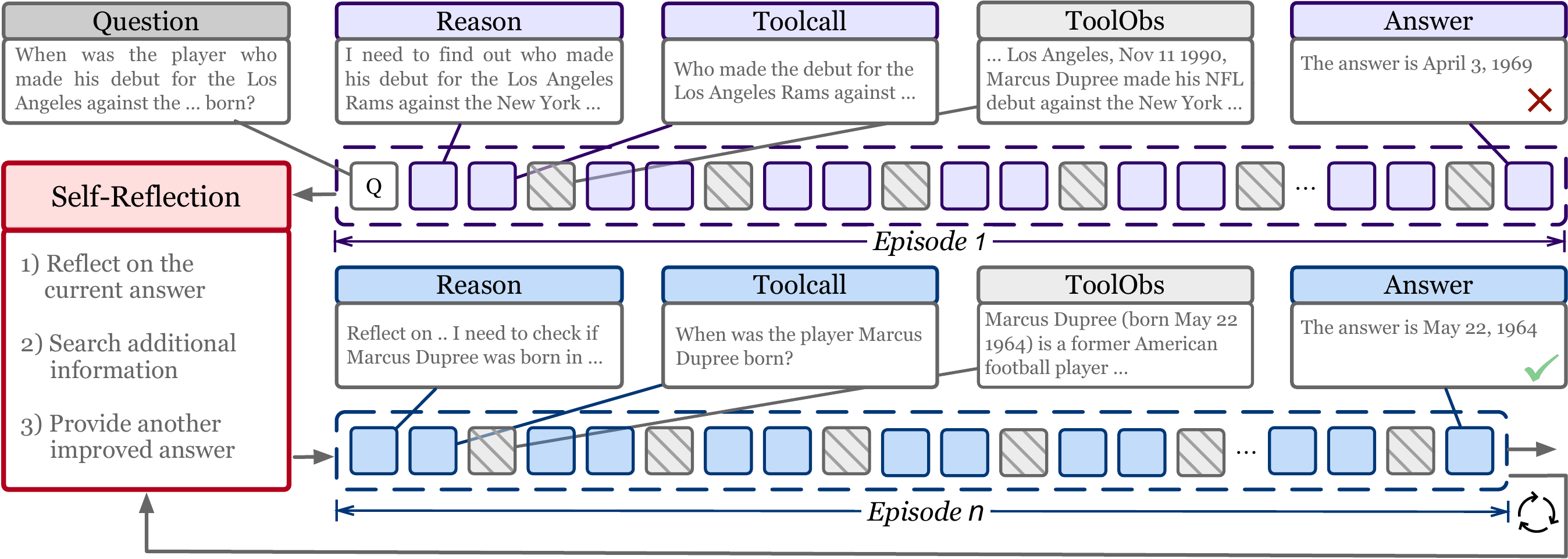}
\caption{An overview of our proposed \method framework. Given a question, the agent first completes an initial episode by interleaving reasoning and tool calls. It then enters an iterative self-reflection loop, where previous episodes serve as experience to inform subsequent searches and answer revisions, enabling iterative improvement across episodes.}
\label{fig:main} 
\vskip -1em
\end{figure}

\subsection{Background} 
\label{sec:background}
Given a dataset $\mathcal{D}$ consisting of question--answer pairs and an external search engine $\mathcal{E}$, our goal is to train a LLM agent to give the answer by iteratively performing reasoning and interacting with $\mathcal{E}$. Many search agent frameworks build upon the ReAct paradigm~\citep{yao2022react,jin2025search}, where agents execute iterative cycles of reasoning and acting until a final answer is reached. When presented with a query, the agent conducts multiple rounds of thought-action-observation sequences. During each round, the  RL-based agent $\pi_{\theta}$ formulates a \textit{internal}  Thought ($\tau$) based on the current context, executes a \textit{external} Action ($\alpha$) such as query, and receives corresponding feedback from search engines as tool Observation ($x$) as shown in Figure~\ref{fig:paradigm}. The interaction trajectory is denoted as:
\begin{align}
a=\left(\tau_0, \alpha_0, x_0, \tau_1, \alpha_1, x_1, \ldots, \tau_{T-1}\right).
\end{align}

The first round contains the prompt, while the final round $\tau_{T-1}$ contains only the thought with the final answer $o$, without any further actions. Given this interaction process, we can directly maximize the RL objective with the final outcome rewards to optimize  the policy:
\begingroup\makeatletter\def\f@size{10}\check@mathfonts\def\maketag@@@#1{\hbox{\m@th\normalfont\normalfont#1}}
\begin{align}
    &\mathcal{J}(\pi_{\theta})=\mathbb{E}_{a\sim \pi_\theta}\big[f_{\text{verifier}}(o,o^*)\big], \label{Eq:RLobjective}
\end{align}
\endgroup
where $o$ denotes the final answer extracted from the completed trajectory $a$, $o^{*}$ is the ground-truth answer, and $f_{\text{verifier}}$ represents either a rule-based or model-based verifier. Although agentic search based on outcome rewards has demonstrated promising performance~\citep{jin2025search,sun2025zerosearch,chen2025learning,shi2025search,wu2025webdancer}, the outcome rewards are  sparse and delayed, leading to ambiguous credit assignment and ineffective search exploration~\citep{liu2021decoupling,feng2025group,wang2025stepsearch}.

\subsection{Meta-RL Framework for Agentic Search}
\label{sec:framework}
In this section, we introduce \method, a Meta-RL framework for agentic search built on a cross-episode meta training scheme. Each meta-episode is modeled as a sequence of episodes, encouraging early exploration and subsequent exploitation of accumulated context, as shown in Figure~\ref{fig:paradigm}. By modeling prior episodes in context through a standardized self-reflection paradigm and propagating this information across episodes, \method enables increasingly informed search, leading to broader exploration and reduced redundancy.

\begin{figure*}[!t]
    \centering
    \includegraphics[width=0.9\linewidth]{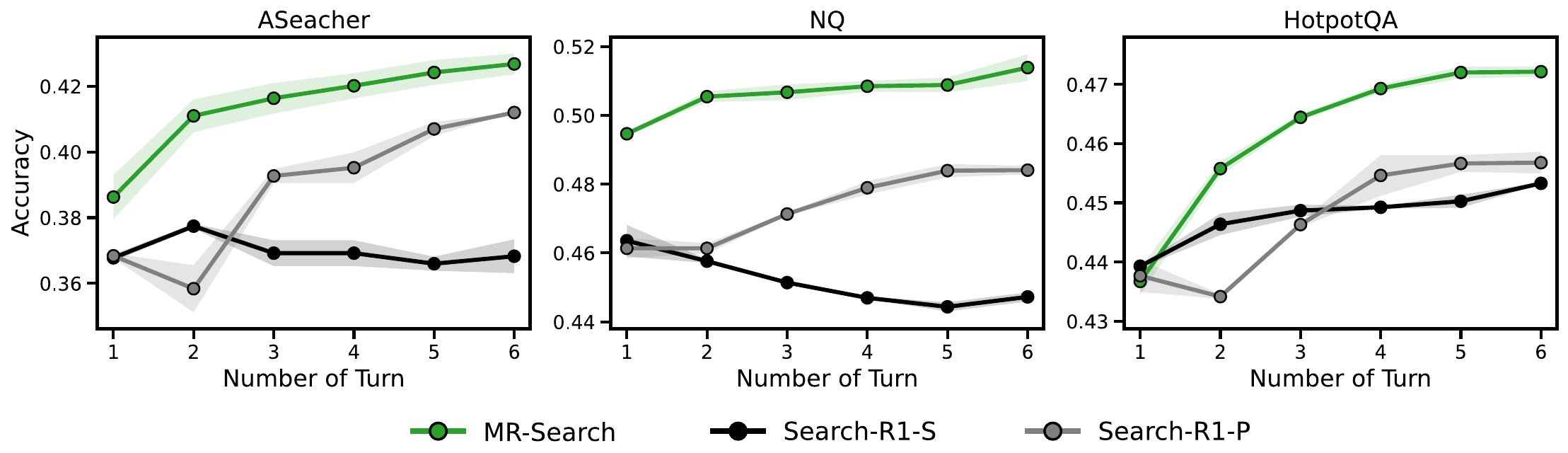}
    \vskip -1em
    \caption{We evaluate \method, Search-R1 with sequential reflection inference (Search-R1-S), and Search-R1 with parallel sampling (Search-R1-P), selecting the most frequent answer among the generated trajectories. Shaded regions
show the standard deviation across 3 runs. We observe that \method achieves the best performance. See \S~\ref{sec:analysis} for details} 
   \vspace{-1em}
    \label{fig:turn_performance_1}
\end{figure*}

\newcommand{\algcmt}[1]{\hfill {\small  #1}}

\begin{wrapfigure}[14]{r}{0.6\columnwidth}
\vspace{-18pt}
\begin{minipage}{0.6\columnwidth}
\begin{algorithm}[H]
\caption{\method: One Sample Update}
\label{alg:metasearch}
\begin{algorithmic}[1]

\State \textbf{Input:} policy $\pi_\theta$, sample $(x,o^*) \sim \mathcal{D}$,
group size $G$, reflection steps $N$
\For{$i = 1, \dots, G$} \algcmt{\Comment{Group of meta-episodes}}

    \State $C \gets x$ \algcmt{\Comment{Initialize context}}

    \For{$n = 0, \dots, N-1$} \algcmt{\Comment{Inner episodes}}
        \State $a_{i,n} \sim \pi_\theta(\cdot \mid C)$
        \State $C \gets C \oplus a_{i,n} \oplus \Call{Reflect}{C}$ \algcmt{\Comment{Self-reflection}}
        \State $r_{i,n} \gets f(a_{i,n}, o^*)$ 
        \algcmt{\Comment{Reward}}
    \EndFor

\EndFor
\State Compute $A_{i,n}$ via Eq.~(7--8) \algcmt{\Comment{RLOO advantage}}
\State $\theta \gets \Call{RL\_Update}{\theta,\{r_{i,n}\}}$ with Eq.~\eqref{Eq:final_objective}


\end{algorithmic}
\end{algorithm}
\end{minipage}
\end{wrapfigure}

The key motivations are twofold: (i) For LLM agents, model limitations often stem from insufficient exploration within a single trajectory rather than inadequate reasoning capacity~\citep{shen2025thinking}; incorrect answers may consistently appear across multiple samples in parallel, as shown by ~\citet{fan2025ssrl, si2026towards}.
(ii) LLMs exhibit strong self-reflection capabilities, enabling inference-time adaptation  that accumulates information from prior interactions across episodes, naturally leveraging in-context learning mechanisms. As shown in Figure~\ref{fig:turn_performance_1}, \method
substantially improves over baselines both with sequential reflection and parallel sampling and its performance grows with more turns, suggesting that the ability for self-reflection emerges as models become stronger in-context meta-learners.

Given an input question, the search agent $\pi_{\theta}$ first generates the first episode, in which it interacts with external tools  until reaching the final answer, following the procedure in \S\ref{sec:background}.
\begin{align}
    {a}_{0}\sim \pi_{\theta}({a}).
\end{align}
Unlike standard RL-based search agents~\citep{jin2025search,song2025r1}, where episodes are independent, \method conditions each episode on the preceding one. After each episode, the search agent invokes the tool to iteratively improve its answer under the self-reflection paradigm. Specifically, given an initial episode, we apply a reflection prompt (Appendix~\ref{appendix:implementation}) that triggers the model to refine its answer by conditioning on previous episodes as context and producing an additional episode with another answer.
\begin{align}
    a_{1}\sim p_{\theta}({a}_{1} \mid {a}_{0}), ~~{a}_{2}\sim p_{\theta}({a}_{2} \mid a_{0},a_{1}) \dots
\end{align}
By repeating the multi-turn self-reflection process  $N-1$ times, in \method, each meta-episode consists of  $N$  episodes sequentially generated by the search agent:
\begin{align}
    y=(a_{0},a_{1}, \cdots,a_{N}),
\end{align}
where we can compute rewards independently for each episode $a$ using the answer $y_{n}$ extracted from the $n$th episode by the verifier. Given this multi-turn process,  we can define the meta-level objective as maximizing the expected reward of the meta-episode:
\begin{align}
    \mathcal{J}_{\text{meta}}({\pi_\theta})=\mathbb{E}_{y \sim \pi_\theta}\Big[\sum\nolimits_{n=0}^{N-1}\gamma^{n} R(s_{n},a_{n})\Big]=\mathbb{E}_{y\sim \pi_\theta}\Big[\sum\nolimits_{n=0}^{N-1}\gamma^{n} f_{\text{verifier}}(o_{n},o^*)\Big],
\end{align}
where $s_{n} = a_{<n}$ denotes the accumulated meta context up to episode $n$, $o_n$ is the answer extracted from the episode $a_{n}$, and $\gamma \in (0,1]$ is the discount factor accounting for future returns. Unless otherwise specified, we set $\gamma = 1$ in this work. In this work, \method conditions on trajectories and explicit reflections from all previous episodes, causing the context length to increase linearly with the number of reflection steps $N$. To mitigate this scalability issue, one can retain only the immediately preceding episode as context, or adopt a context management protocol that summarizes prior episodes before carrying them forward to subsequent episodes. In \S\ref{exp:extension}, we empirically find that only keeping the immediately preceding episode as context also also works well for \method.

\subsection{Policy Optimization with Multi-Turn Advantages}
\label{sec:optimization}
To optimize the policy, instead of estimating advantages using separate value functions as in PPO~\citep{schulman2017proximal}, we propose an approach that incorporates turn-level reward signals while maintaining unbiased policy optimization. Specifically, for each question, we sample a group of $G$ meta-episodes, $\mathcal{G}=\{y_i\}_{i=1}^G$. To make rewards comparable across episodes and reduce variance, we aggregate rewards at the same episode over meta-episodes and apply Leave-One-Out  (RLOO) estimation~\citep{kool2019buy,ahmadian2024back}:
\begin{align}
    \tilde{r}_{i,n}={r}(s_{i,n},a_{i,n})-\text{mean}_{j \neq i}{r}(s_{j,n},s_{j,n})={r}(s_{i,n},a_{i,n})-\frac{1}{G-1} \sum\nolimits_{j \neq i} {r}(s_{j,n},a_{j,n}),
\end{align}
where $\tilde{r}_{i,n}$ is the reward of the $n$th episode in the $i$th meta-episode (i.e., $y_{i,n}$). Compared to GRPO~\citep{shao2024deepseekmath}, RLOO provides an unbiased advantage estimation~\citep{bereket2025uncalibrated}. While $\tilde{r}_{i,n}$ captures the relative quality of each episode, it reflects immediate effects and ignores the impact of future. To include long-horizon dependencies, we compute a discounted cumulative advantage to propagate rewards backward to earlier turns:
\begin{align}
   A_{i,n}=\sum\nolimits_{n'=n}^N \gamma^{n'-n} \tilde{r}_{i,n'}. \label{Eq:adv}
\end{align}
As the baseline used in RLOO, $\text{mean}_{j \neq i}{r}(s_{j,n},s_{j,n})$, does not depend on the current action ${y}_{i,n}$, it provides an unbiased estimate of the turn-level advantage~\citep{sutton1999policy}. 

With the discounted turn-level advantages above, we optimize the policy using a clipped surrogate off-policy objective in PPO~\citep{schulman2017proximal}. Formally, the objective is: 

\begingroup\makeatletter\def\f@size{9.5}\check@mathfonts\def\maketag@@@#1{\hbox{\m@th\normalfont\normalfont#1}}
\begin{align}
\frac{1}{G}
  \sum_{i=1}^{G} \frac{1}{|{y}_i|}\sum_{n=1}^{|{y}_i|}
  \min \Bigl(
  \frac{\pi\left({y}_{i,n} \mid x, y_{i,<n};\theta\right)}{\pi(y_{i,n} \mid x, y_{i, <t}; \overline{\theta})}\, A_{i,n},\
    \operatorname{clip} \bigl(
   \frac{\pi\left(y_{i, n} \mid x, y_{i,<n};\theta\right)}{\pi(y_{i,n} \mid x, y_{i, <n};\overline{\theta})},\;
      1 - \varepsilon_{},\;
      1 + \varepsilon_{}
    \bigr)\,
   A_{i,n}
  \Bigr), \label{Eq:final_objective}
\end{align}
\endgroup
where $\pi(y_{i,n} \mid x, y_{i,<n}; \theta)$ and $\pi(y_{i,n} \mid x, y_{i,<n}; \overline{\theta})$ denote the current and old policy models over the steps, and we broadcast each step’s advantage signal to all tokens in that step~\citep{shao2024deepseekmath}. $\epsilon$ is the is the clipping ratio. We additionally mask out tool output
tokens from the loss, following previous work~\citep{jin2025search}. Optimizing the policy with the above objective enables the policy to capture both global trajectory quality and local step effectiveness. 
In \S\ref{ablation}, we compare our objective with PPO~\citep{schulman2017proximal} and MT-GRPO~\citep{zeng2025reinforcing} using our designed process rewards. The results show that our objective consistently achieves better performance under the meta-RL framework. The full algorithm of \method (one-sample training) is summarized in Algorithm~\ref{alg:metasearch}.

\subsection{Discussion}
\label{sec:discussion}

\textbf{Exploration \& Exploitation}. 
As discussed above, \method leverages experience from previous episodes to guide subsequent exploration. By default, all episodes contribute rewards according to Eq.~\eqref{Eq:adv}. To promote unstructured exploration, we can optionally mask rewards for designated exploration episodes while retaining rewards for exploitation episodes~\citep{stadie2018some}. 
Exploration episodes are fully visible during the forward pass but receive zero reward during backpropagation, so gradients are driven only by exploitation episodes. Concretely, the advantage is computed using a masked return:
\begin{align}
A_{i,n}=\sum\nolimits_{n'=n}^N \gamma^{n'-n}\tilde{r}_{i,n'}m_{n'},
\end{align}
where $m_{n'} \in \{0,1\}$ indicates exploitation (1) or exploration (0).  The policy gradient is therefore computed using this masked return rather than the standard discounted return. By zeroing out exploration rewards, we encourage the policy to prioritize long-term gains from improved context adaptation rather than short-term episode feedback. Although exploration episodes do not directly contribute to the gradient, they serve as contextual adaptation steps that improve environment identification and lead to higher rewards in subsequent exploitation episodes. We provide empirical analysis of this strategy in \S\ref{exp:extension} and find that it is helpful for ASearcher, which requires multi-turn search.

\textbf{Meta-RL at the Step Level}. 
While \method models agentic search as a meta-episode composed of multiple reflection episodes, it treats each full trajectory with an answer as a single optimization unit. However, in long-horizon reasoning, inefficiencies often arise at a finer granularity, such as individual tool calls or intermediate reasoning steps. The same principle of \method extends naturally to semantically meaningful sub-episodes. For instance, in agentic search, each tool-interaction step can be treated as a micro-episode. During training, we prompt the model to produce an intermediate answer after each tool call as shown in Figure~\ref{fig:step} in Appendix and evaluate it with the verifier to obtain step-level rewards. Converting these substructures into micro-episodes enables localized credit assignment, transforming long trajectories into  reflection steps. This dense supervision promotes informative intermediate reasoning and reduces redundant exploration. We empirically show that this extension also achieves strong performance (\S\ref{exp:extension}).

%% file: sections/experiments.tex
\section{Experiments}\label{sec:exps}

\subsection{Experimental Setup}

\textbf{Datasets} We conduct evaluations on the following datasets:
\textit{(1) General Question Answering}: NQ~\citep{kwiatkowski2019natural}, TriviaQA~\citep{joshi2017triviaqa}, and PopQA~\citep{mallen2022not}; and
\textit{(2) Multi-Hop Question Answering}: HotpotQA~\citep{yang2018hotpotqa}, 2WikiMultiHopQA~\citep{ho2020constructing}, Musique~\citep{trivedi2022musique}, and Bamboogle~\citep{press2022measuring}. For training, we merge the NQ and HotpotQA training sets to construct a unified dataset for all finetuning approaches, following the setup in~\citep{jin2025searchr}. For evaluation, we use the test or development splits of the seven datasets listed above. We additionally include a synthetic dataset, ASearcher~\citep{gao2025beyond}, which is more complex than NQ/HotpotQA and requires long multi-turn search. We split ASearcher into 90\% training and 10\% evaluation sets. We split ASearcher into 90\% training and 10\% evaluation sets. Detailed dataset descriptions and statistics are provided in Appendix~\ref{appendix:dataset}.

\begin{table*}[!t]
    \centering
    \caption{Main accuracy (\%) on search-based QA benchmarks. The best results are marked in \textbf{boldface}. We compare with baselines that rely on outcome rewards (ReSearch and Search-R1) and those that use process rewards with external models (PPRM and StepResearch).
    }
    \label{tab:results}
    \setlength{\tabcolsep}{1.9mm}
    \resizebox{0.99\linewidth}{!}{
    \begin{tabular}{p{2mm}|lcccccccc}
        \toprule
        \multicolumn{1}{p{3mm}}{} & \multicolumn{1}{l}{\multirow{2}{*}[-.3em]{\textbf{Method}}} & \multicolumn{3}{c}{\textbf{Single-Hop QA}} & \multicolumn{4}{c}{\textbf{Multi-Hop QA}}  \\
        \cmidrule(lr){3-5} \cmidrule(lr){6-9}
        \multicolumn{1}{p{3mm}}{} & &  NQ &  TriviaQA & PopQA &  HotpotQA & 2wiki & Musique &  Bamboogle &  Avg. \\
        \midrule
        \multirow{8}{*}{\rotatebox[origin=c]{90}{\textbf{Qwen2.5-3b}}}
        & Direct Inference & 10.6 & 28.8 & 10.8 & 14.9 & 24.4 & 2.0 & 2.4 & 13.4 \\
        & Search-o1 & 23.8 & 47.2 & 26.2 &  22.1 & 21.8 & 5.4 & 32.0 & 25.5 \\
        & ReSearch & 42.7 & 59.7 & 43.0 &  30.5 & 27.2  & 7.4 & 11.5 & 30.4 \\
        &  Search-R1 & 46.2 & 62.2 & 45.6 & 32.6 & 31.0 & 7.7 & 17.6 & 34.7 \\
        \cdashline{2-10}
        \addlinespace[0.8ex]
        & PPRM & 42.3 & 56.5 & 41.1 & 35.3 & 34.0 & 12.7 & 28.0 & 35.7 \\
        & StepResearch & 44.6 & 61.5 & 45.6 &  37.3 & 33.8 & 10.5 & 32.5 & 38.0\\
        \cdashline{2-10}
         \addlinespace[0.8ex]
        & \method & \textbf{47.7} & \textbf{63.5} & \textbf{46.0} & \textbf{41.9} &  \textbf{40.1} &  \textbf{16.5} &  \textbf{34.4} &  \textbf{41.4} \\
        \midrule
        \multirow{8}{*}{\rotatebox[origin=c]{90}{\textbf{Qwen2.5-7b}}}
        & Direct Inference & 13.4 & 40.8 & 14.0 &  18.3 & 25.0 & 3.1 & 12.0 & 18.1 \\
        & Search-o1 & 15.1 & 44.3 & 13.1 &  18.7 & 17.6 & 5.8 & 29.6 & 20.6\\
        & ReSearch & 36.6 & 60.5 & 39.1&  37.8 & 38.6 & 16.6 & 37.6 & 38.1 \\
        & Search-R1 & 45.9 & 63.2 & 44.9 & 43.9 & 38.7 & 18.1 & 40.0& 42.1 \\
        \cdashline{2-10}
       \addlinespace[0.8ex]
        & PPRM & 45.8 & 61.0 & 43.7 & 38.6 & 35.5 & 14.7 & 35.5 & 39.3 \\
        & StepResearch & 47.3 & 63.6 & 43.1 &  43.9 & 41.8 & 20.5 & 43.5 & 43.4 \\
        \cdashline{2-10}
         \addlinespace[0.8ex]
        & \method &  \textbf{50.2} &  \textbf{66.6} &  \textbf{47.2} &  \textbf{46.8} &  \textbf{43.6} &  \textbf{22.1} &  \textbf{45.2} &  \textbf{46.0} \\ 
        \bottomrule
    \end{tabular}
    }
\end{table*}

\textbf{Evaluation Metrics}
For evaluation metrics, we follow~\citet{jin2025searchr}: we first normalize both the predicted and ground-truth answers, and then compute Exact Match (EM) score. 
EM achieves true if and only if the predicted answer exactly matches any ground-truth answer. For all methods, we sample a single trajectory per question and report the average EM for the last valid prediction for questions following~\citet{jin2025search}.
\begin{figure*}[!t]
    \centering
    \includegraphics[width=0.98\linewidth]{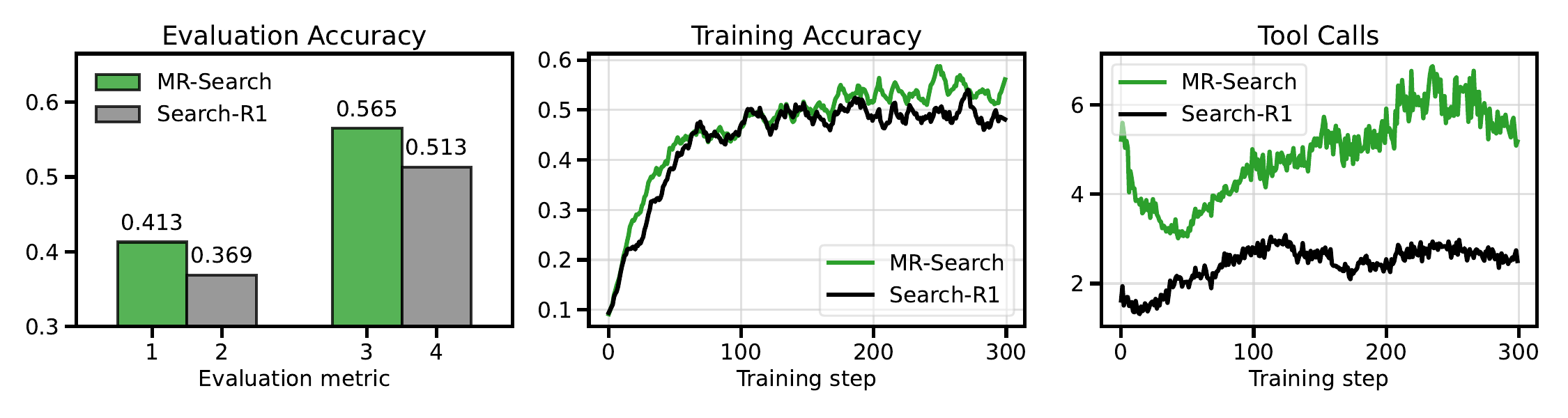}
    \vskip -1em
    \caption{Test performance, training curves of reward and search frequency on ASearcher, evaluated with Qwen2.5-7B-Base.  Additional results are provided in Appendix~\ref{appendix:training_dynamic}.} 
    \label{fig:as_performance}
        \vskip -1em
\end{figure*}

\textbf{Setup} 
We mainly conduct experiments using Qwen-series models~\citep{yang2024qwen2} (Qwen2.5-3B-Base and Qwen2.5-7B-Base). Following~\citep{jin2025searchr}, we use the 2018 Wikipedia dump~\citep{karpukhin2020dense} as the knowledge source and E5~\citep{wang2022text} as the retriever. We fix the number of retrieved documents to three across all methods for a fair comparison. Unless otherwise specified, we report results at turn 3 for our models.
The detailed implementation settings are provided in Appendix~\ref{appendix:implementation}.


\textbf{Baselines}
To evaluate the effectiveness of \method, we compare it against the following recent baselines:
(1) Inference without finetuning: methods that directly use the base model, including direct inference without retrieval and Search-o1~\citep{li2025search} with retrieval.
(2) Finetuning-based methods that learn a policy to integrate the search tool without step-level supervision, including ReSearch~\citep{chen2025learning} and Search-R1~\citep{jin2025search}.
(3) Finetuning-based methods that learn a policy to integrate the search tool with step-level supervision, including PPRM~\citep{anonymous2025preferencebased} and StepResearch~\citep{wang2025stepsearch}. We also compare with Search-R1 with our multi-turn reflection mechanism during inference. in \S~\ref{sec:analysis}. The detailed description of baselines are given in Appendix~\ref{appendix:baselines}.

\subsection{Main Results}
Table~\ref{tab:results} summarizes main results on benchmarks.  Among the approaches, our \method achieves a substantial margin over GRPO with outcome rewards (Search-R1), yielding 9.2\% and 19.3\% relative improvements on average for the Qwen2.5-7B-Base and Qwen2.5-3B-Base, respectively. This highlights the significant benefits of designing process rewards for agentic search. Remarkably, \method remains highly effective on the small Qwen2.5-3B model, whereas RL methods that rely only on sparse outcome rewards struggle to elicit multi-turn search behavior for good performance. Compared to other methods that rely on external models to obtain the process reward such as StepResearch and PPRM, our \method achieves better performance. This confirms that agentic search can benefit greatly from our designed free process rewards, and that \method can effectively leverage this process supervision to achieve better performance. 
Figure~\ref{fig:as_performance} shows the results on the ASearcher datasets. Compared to Multi-Hop QA, ASearcher requires longer-horizon, multi-turn search~\citep{gao2025beyond}. From the results, we observe that \method significantly outperforms Search-R1, achieving 10.2\% and 9.5\% relative improvements in EM and F1, respectively, demonstrating the effectiveness of  \method on complex ASearcher tasks.

\textbf{Ablation Study.}
\label{ablation}
We study the effects of the key design choices of \method. Specifically, we consider the following ablations: (i) we compare our optimization algorithm (Section~\S\ref{sec:optimization}) with PPO~\citep{schulman2017proximal} and MT-GRPO~\citep{zeng2025reinforcing} within our Meta-RL framework; (ii) we set the discount factor $\gamma=0$, which removes future credit assignment. From the results in Table~\ref{tab:rl-ablation}, we observe that our proposed multi-turn RL algorithm consistently outperforms PPO and MT-GRPO when trained with episode turn-level rewards, demonstrating the effectiveness of our optimization strategy in leveraging dense feedback from reflection. Moreover, both PPO and MT-GRPO underperform GRPO with outcome rewards (Search-R1) on single-hop NQ and PopQA, whereas our method does not, indicating stronger generalization and robustness. We further observe that removing the discount factor substantially degrades performance and causes the training process to converge to poor local optima. As discussed above, a possible explanation is that an incorrect episode does not necessarily imply that intermediate episodes are uninformative.

\begin{figure*}[!t]
    \centering
    \includegraphics[width=0.98\linewidth]{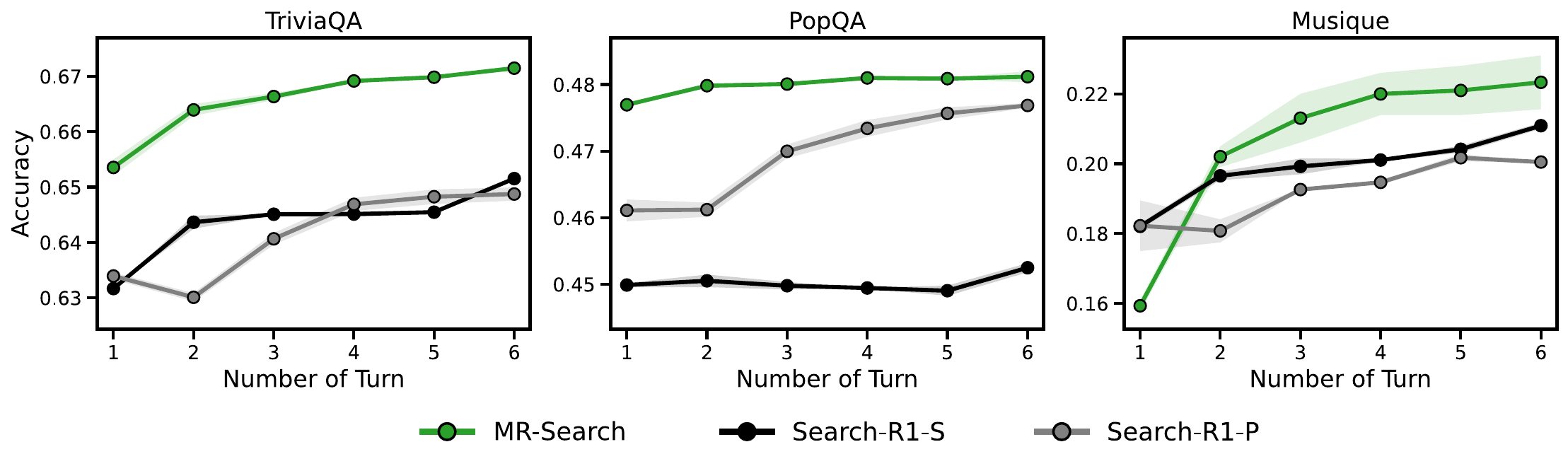}
    \vskip -1em
    \caption{\method, Search-R1 with sequential reflection turns (Search-R1-S), and Search-R1 with parallel sampling (Search-R1-P), selecting the most frequent answer among them.} 
        \vskip -1em
    \label{fig:turn_performance}
\end{figure*}

\begin{table*}[!t]
\centering
\caption{Ablation of discount factor and training algorithm evaluated on Qwen2.5-7B-Base.}
\label{tab:rl-ablation}
\setlength{\tabcolsep}{1.8mm}
\resizebox{0.9\linewidth}{!}{
\begin{tabular}{lcccccccc}
\toprule
\multirow{2}{*}{\raisebox{-0.4em}{\textbf{Method}}}
& \multicolumn{3}{c}{\textbf{Single-Hop QA}} 
& \multicolumn{4}{c}{\textbf{Multi-Hop QA}} \\
\cmidrule(lr){2-4} \cmidrule(lr){5-8}
&  NQ & TriviaQA & PopQA 
& HotpotQA & 2wiki & Musique & Bamboogle & Avg.
\\
\midrule
ReSearch & 36.6 & 60.5 & 39.1&  37.8 & 38.6 & 16.6 & 37.6 & 38.1 \\
 Search-R1 
& 46.4 & 64.1  & 44.8
& 43.0 & 42.5 & 19.5 & 44.0 & 43.5
 \\
  \midrule
\method w. $\gamma=0$ 
& 49.3 & 63.9 & 45.6
& 44.3 & 41.7 & 19.3 & 42.6 & 43.8
 \\
 \midrule
\method w. PPO 
& 43.9 & 63.7 & 42.5
& 41.3 & 40.5 & 18.6 & 43.5 & 42.0 
 \\
\method w. MT-GRPO
& 46.1 & 64.5 & 44.5
& 44.7 & \textbf{44.2} & 20.7 & 45.1 & 44.3
\\
 \midrule
\method 
& \textbf{50.2} & \textbf{66.6} & \textbf{47.2}
& \textbf{46.8} & {43.6} & \textbf{22.1} & \textbf{45.2} & \textbf{46.0}
\\
\bottomrule
\end{tabular}
}
\vskip -2em
\end{table*}

\subsection{Further Analysis}
\label{sec:analysis}
\paragraph{Training Dynamics.} 
For a more comprehensive understanding of \method, we visualize its training dynamics. Figure~\ref{fig:as_performance} shows the training reward dynamics and  of Search-R1 and \method. We observe that \method exhibits stable convergence during training and consistently achieves higher training reward than Search-R1. This indicates that \method effectively leverages iterative reflection to refine answers, leading to progressively improved final answers throughout training of multi-turn RL training. We can observe that \method calls the search engine more frequently than Search-R1, demonstrating that \method can dynamically adjust the number of search calls according to the complexity of tasks.

 \begin{table*}[!t]
    \centering
    \caption{Comparison with variants of \method: encouraging exploration, \method at the step level and context management (keeping one preceding episode as context).}
    \label{tab:app-variants}
    \resizebox{1\linewidth}{!}{
    \begin{tabular}{lcccccccc}
        \toprule
         \multicolumn{1}{l}{\multirow{1}{*}{\textbf{Method}}} 
         &  NQ &  TriviaQA & PopQA &  HotpotQA & 2wiki & Musique &  Bamboogle & ASearcher  \\
         \midrule
        Search-R1 & 45.9 & 63.2 & 44.9 & 43.9 & 38.7 & 18.1 & 40.0& 36.9 \\
        \midrule
        \method &  \textbf{50.2} &  \textbf{66.6} &  \textbf{47.2} &  \textbf{46.8} &  \textbf{43.6} &  \textbf{22.1} &  \textbf{45.2} &  41.3 \\ 
        \midrule
        \method Exploration & 48.3 & 65.1 & 46.4 & 44.7 &  39.4 &  21.8 &  44.0 &  \textbf{43.2} \\       
        \method Step Level&  48.6 & 64.6 & 45.7 & 42.3 & 41.4 & 16.3 & 41.6 & 38.4  \\
        \method Short Context &  48.1 & 65.9 & 45.2 & 44.6 & 41.0 & 19.3 & 47.2 &  40.5  \\
        \bottomrule
    \end{tabular}
    }
    \vskip -1em
\end{table*}

\textbf{Test-time Scaling.}
We also evaluate how \method scales with additional reflection turns at test time. We extrapolate the number of reflection turns beyond
training (3 turns) by appending the entire interaction history to the context at each turn. As shown in Figures~\ref{fig:turn_performance_1} and \ref{fig:turn_performance}, the single-turn method Search-R1 with our reflection mechanism  yields only marginal gains when additional reflection turns are allowed, since its training objective is optimized for a single turn. However, when additional reflection turns are allowed,  \method achieves significantly higher performance, exhibiting the steep improvement curve. These results suggest that multi-turn reflection with \method enhances the model’s ability to iteratively refine and optimize its search across turns and enables effective extrapolation.

\textbf{Case Study.}
We present inference cases of models trained with \method. As illustrated by the examples in Appendix~\ref{appendix:case_study}, the model is able to execute multi-turn agentic tasks through iterative tool calls and autonomous information aggregation. During subsequent reasoning episode, the model revisits and revises  answers in light of newly acquired evidence, ultimately producing correct final answers. These observations highlight the benefits of self-reflection for agentic search. Moreover, when the answer is already accurate as shown in Cases 2 and 3, the model can preserve it and avoid unnecessary revisions, demonstrating its ability to selectively refine reasoning based on retrieved information.

\subsection{Extensions} 
\label{exp:extension}
In this section, we conduct preliminary experiments to investigate the effects of exploration vs. exploitation, step-level meta-RL (discussed in \S~\ref{sec:discussion}), and context management (i.e., retaining only the immediately preceding episode as context). To study exploration and exploitation, we designate the first two episodes as exploration and the last two as exploitation. Table~\ref{tab:app-variants} in provides detailed results. From Table~\ref{tab:app-variants}, we observe that all variants significantly outperform the baseline Search-R1, demonstrating the effectiveness of these variants. Moreover, \method with step-level meta-RL achieves a substantial improvement over GRPO with outcome rewards (Search-R1). This result suggests that agentic search can benefit from process-level rewards under meta-RL training, and that \method effectively leverages such process supervision to achieve better performance. Furthermore, as shown in Table~\ref{tab:app-variants}, encouraging exploration by assigning rewards to the first episode is beneficial for the more complex ASearcher dataset, which requires more interaction with tools.

%% file: sections/conclusion.tex
\begin{section}{Conclusions}\label{sec:conclusion}
We study agentic search under sparse outcome rewards and propose \method, an in-context meta-reinforcement learning framework that enables structured cross-episode exploration via explicit self-reflection. By conditioning each episode on prior trajectories and reflections, \method transforms independent search attempts into a progressively informed search process, improving exploration without relying on external process reward models. To support this multi-turn reflective setting, we introduce a turn-level grouped advantage formulation that provides unbiased and fine-grained credit assignment while remaining critic-free. Extensive experiments across diverse benchmarks show that \method consistently outperforms outcome-only RL baselines. Overall, our results highlight the importance of in-context meta-learning for effective agentic reinforcement learning of LLMs.
\end{section}

%% file: sections/appendix.tex
\subsection{Experimental Details}
\subsubsection{The Details of Datasets}
\label{appendix:dataset}

\paragraph{NQ}~\citep{kwiatkowski2019natural}: NQ is question answering (QA) data set. Questions consist of real anonymized, aggregated queries issued
to the Google search engine. The training and test sets contain 79,168 and 3,610 samples, respectively.

\paragraph{TriviaQA}~\citep{joshi2017triviaqa}: TriviaQA is a challenging reading comprehension dataset containing over question-answer-evidence triples, which provide high quality distant supervision for answering the questions. The test sets we used contain 11,313 samples.

\paragraph{PopQA}~\citep{mallen2022not}: PopQA is a large-scale open-domain question answering dataset, consisting of 14,267 entity-centric QA pairs. Each question is created by converting a knowledge tuple retrieved from Wikidata using a template. 

\paragraph{HotpotQA}~\citep{yang2018hotpotqa}: HotpotQA is a large-scale multi-hop QA benchmark featuring  Wikipedia-based Q\&A pairs with sentence-level supporting evidence. The training and test sets contain 90,447 and 7,405 samples, respectively.

\paragraph{2WikiMultiHopQA}~\citep{ho2020constructing}: 2WikiMultiHopQA is a multi-hop question answering dataset designed to more reliably test a model’s inference across multiple pieces of evidence. The test set contains 7,405 QA pairs.

\paragraph{Musique}~\citep{trivedi2022musique}: MuSique is a multihop question answering dataset constructed to enforce genuine multi-step reasoning. The test set contains 2,417 QA pairs.

\paragraph{Bamboogle}~\citep{press2022measuring}. Bamboogle is a dataset with multi-hop questions, where all questions are
sufficiently difficult to be unanswerable by a popular internet search engine, but where both supporting pieces of evidence can be found in Wikipedia. Bamboogle contains 125 test QA pairs.

\paragraph{ASearcher}~\citep{gao2025beyond}. ASearcher is a synthetic multi-turn dataset whose synthesis pipeline is largely based on Wikipedia. We use the preprocessed version from~\citep{cut2025}, which applies a three-step filtering pipeline, including the removal of Chinese and math samples as well as rejection sampling, resulting in 14k samples. We split ASearcher into 90\% training and 10\% evaluation sets to decouple the effects of data distribution.

\begin{figure*}
\begin{tcolorbox}[colback=red!6!white,colframe=black, boxsep=0pt,top=8pt,bottom=8pt,left=8pt,right=8pt]
\paragraph{Generation Prompt} Answer the given question. You must conduct reasoning inside \verb|<think>| and \verb|</think>| first every time you get new information. After reasoning, you must call a search engine by \verb|<search>| query \verb|</search>|, and it will return the top search results between \verb|<information>| and \verb|</information>|. After every time you get new information, you must try to provide the answer inside \verb|<answer>| and \verb|</answer>| without detailed illustrations. For example, \verb|<answer>| xxxx \verb|</answer>|. Question: \{question\}
\end{tcolorbox}
\label{fig:gen_prompt}
\end{figure*}

\subsubsection{The Details of Baselines}
\label{appendix:baselines}
We provide detailed descriptions of baselines. We consider three types of baselines: Inference without fine-tuning, Fine-tuning-based methods without step-level supervision and Fine-tuning-based methods with step-level supervision.

\textbf{Search-o1}~\citep{li2025search} is the search-enhanced reasoning framework, which integrates the agentic RAG mechanism and reason-in-document module.

\textbf{ReSearch}~\citep{chen2025learning} is an RL-based framework that trains LLMs to interleave reasoning with explicit search actions, deciding when and how to query and then using retrieved evidence to continue multi-hop reasoning.

\textbf{Search-R1}~\citep{jin2025search} extends RL-based reasoning by enabling LLMs to autonomously generate search queries during multi-turn reasoning. 

\textbf{PPRM}~\citep{anonymous2025preferencebased} is a principle process reward model that provides step-wise signals to guide GRPO-based RL.

\textbf{StepResearch}~\citep{wang2025stepsearch} trains search agents with step-wise PPO using intermediate rewards and token-level supervision to better guide multi-hop retrieval and reasoning.

\begin{figure*}
\begin{tcolorbox}[colback=blue!6!white,colframe=black,boxsep=0pt,top=8pt,bottom=8pt,left=8pt,right=8pt]
\paragraph{Reflection Prompt} Reflect on your current answer to the question and provide another answer by searching for additional external information using search engines. You must conduct reasoning inside \verb|<think>| and \verb|</think>|  first every time you get new information. After reasoning, if you find you lack some knowledge, you can call a search engine by \verb|<search>| query \verb|</search>| and it will return the top searched results between \verb|<information>| and \verb|</information>|. You can search as many times as your want. If you find no further external knowledge needed, you can directly provide the answer inside \verb|<answer>| and \verb|</answer>|., without detailed illustrations.
\end{tcolorbox}
\label{fig:reflect_prompt}
\end{figure*}

\subsubsection{The Details of Implementation}
\label{appendix:implementation}
Our implementation is built upon Search-R1 based on VeRL. The generation prompt and reflection Prompt are given in the colored box.

\textbf{Hyperparameters}. 
For training, we use AdamW~\citep{loshchilov2017decoupled} as the optimizer and set the learning rate to 1e-6 without warmup. The top-p and temperature for rollout are both set to 1. The total number of training steps is 300. The number of documents returned by the retrieval engine is 3. The group size for advantage calculation is set to 5. The context length is set to 8K and 16K for the NQ/HotpotQA and Asearcher datasets, respectively. We also set the maximum number of tool calls in each episode to 3 and 5 for the NQ/HotpotQA and Asearcher datasets, respectively.
For evaluation, all other settings (e.g., the retriever configuration) are kept the same as in training; 
the only difference is that we disable sampling and use greedy decoding with $\texttt{temperature}=0$, and $\texttt{top\_p}=1.0$.

\paragraph{Computation Resources}
All RL training is performed on $8 \times$ NVIDIA Tesla H100 (80GB) GPUs, and we use an additional 
$2 \times$ NVIDIA Tesla H100 (80GB) GPUs to serve the retriever.

\subsection{Additional Experimental Results}
\label{appendix:additional_results}
\subsubsection{More Results of Training Dynamics}
\label{appendix:training_dynamic}
\begin{figure*}[!t]
    \centering
    \includegraphics[width=1\linewidth]{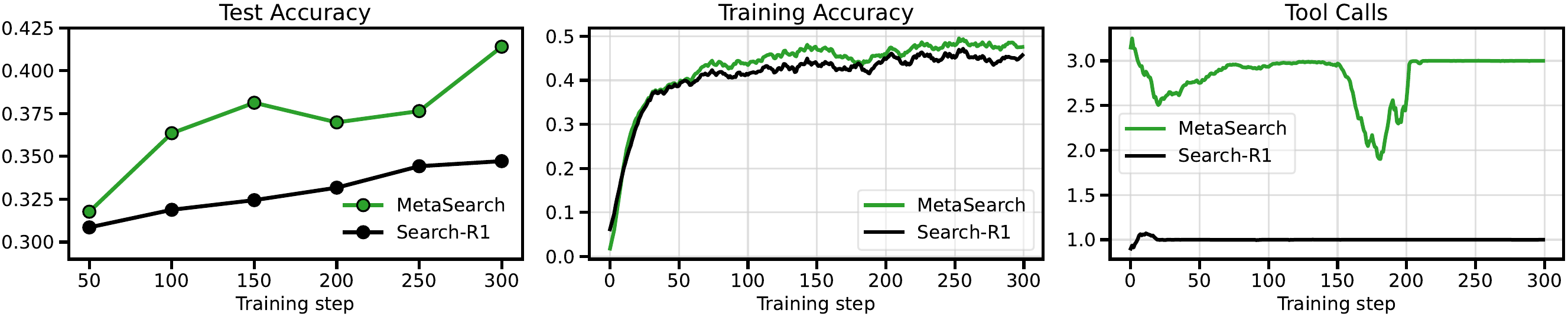}
\caption{The training dynamics of \method and Search-R1 in terms of test accuracy, training accuracy, and the number of tool calls on Qwen2.5-3B-Base.}
    \label{fig:step3B}
    \includegraphics[width=1\linewidth]{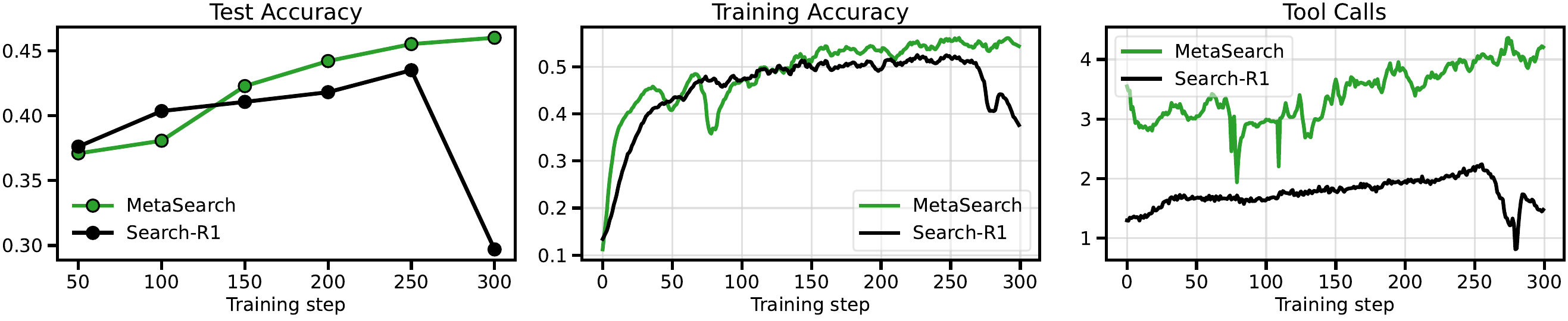}
\caption{The training dynamics of \method and Search-R1 in terms of test accuracy, training accuracy, and the number of tool calls on Qwen2.5-7B-Base.}
    \label{fig:step7B}
\end{figure*}

In Figures~\ref{fig:step3B} and \ref{fig:step7B}, we further visualize the training dynamics of \method. We can observe that \method exhibits more stable convergence during training and consistently achieves higher reward than Search-R1 baseline. These results indicate that the process rewards at episode level with self-reflection introduced by \method provide more informative and reliable learning signals, leading to improved stability and training effectiveness.


\begin{figure*}[!t]
    \centering
    \includegraphics[width=0.98\linewidth]{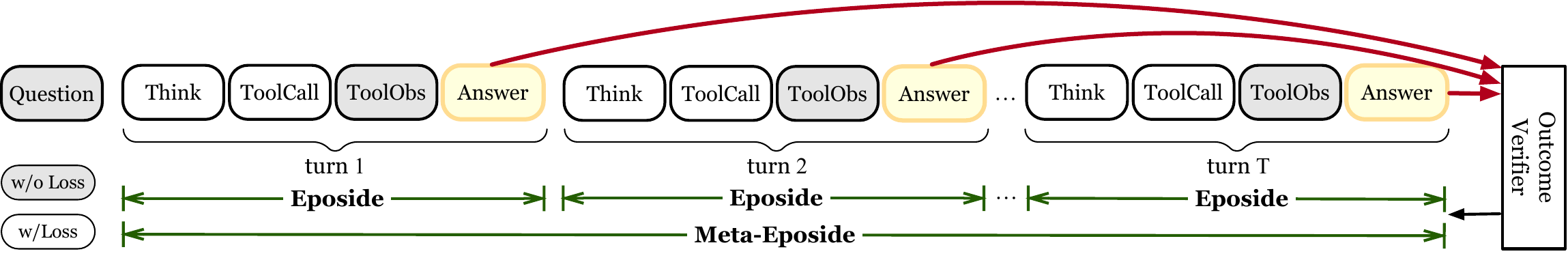}
    \caption{The idea of \method can be extended beyond full trajectories to any semantically meaningful episode that can be explicitly segmented. For example, in agentic search, each tool-interaction step can naturally form an episode. We can decompose the one interaction process into multiple self-reflective trials, where each tool-interaction step contributes a candidate answer to the original query via prompting. During training, at each interaction step, we monitor changes in the agent’s accuracy with respect to the target and use this signal as a dense training objective to promote effective exploration.} 
    \label{fig:step}
\end{figure*}


\subsubsection{Case Study}
\label{appendix:case_study}
In Tables~\ref{tab:case}-\ref{tab:case4}, we provide some inference trajectory of the model trained by \method.

\begin{table*}[h]
\centering
\footnotesize
\caption{Case study 1: An
early intermediate answer is incorrect due to ambiguous retrieval results, but is subsequently reconsidered through additional search and self-reflection, leading to the correct final answer.}\label{tab:case}
\begin{tabular}{p{13.5cm}}
\toprule
\textbf{Question}: During a major geopolitical event in the early Cold War, Robert Menzies exploited rising tensions to portray the Labor Party as 'soft on Communism,' leading to the defeat of a Prime Minister who attended a religious school. In the same period, which British maritime patrol aircraft, known for its role in anti-submarine warfare and derived from the world's first jet airliner, was frequently deployed to a U.S. military base in the Atlantic to support anti-submarine operations during the latter stages of the Cold War? \\
\midrule
\textbf{Ground Truth}: 2002 \\
\midrule
\textbf{\method }: \\
\think{To determine the year in which the Victorian government department responsible for agriculture and energy policy was established, we need more information about the department and its merger with another department in 2013.} \\
\search{Query: Victorian government department responsible for agriculture and energy policy merged with another department administering conservation strategy involving biodiversity offsets in 2013. When was this department established?} \\
\info{Document (Title: Department of the Environment (Australia, 2013–16)) by way of an Administrative Arrangements Order issued on 18 September 2013. It absorbed the responsibilities of the former Department of Sustainability, Environment, Water, Population and Communities (DSWEPaC) and climate change from the former Department of Industry, Innovation, Climate Change, Science, Research and Tertiary Education. The department was dissolved in July 2016 and its functions, along with energy policy functions, were moved to the newly established Department of the Environment and Energy. From when it...} \\
\think{It seems the search did not directly provide the information. Let's try another search with a specific focus on the department related to agriculture and energy policy:}\\
\search{Query: When was the Department of Economic Development, Jobs, Transport and Resources established in Victoria?} \\
\info{Document (Title: Department of Economic Development, Jobs, Transport and Resources) Department of Economic Development, Jobs, Transport and Resources The Department of Economic Development, Jobs, Transport and Resources (DEDJTR) is a department of the Government of Victoria. It was created on 1 January 2015 by the government of Premier Daniel Andrews when the number of government departments was reduced from 9 to 7, and assumed responsibility for ministerial portfolios previously spread across 5 departments. The department was established after the 2014 Victorian election by the new Labor government, using the powers of the Premier of Victoria and the Governor-in-council under the Public ...}\\
\think{This information indicates that the Department of Economic Development, Jobs, Transport and Resources was established on 1 January 2015. Therefore, the year in which the Victorian government department responsible for agriculture and energy policy was established is:} \\
\answer{2015} \\
\prompt{Reflect on your current answer to the question and provide an improved answer by searching for additional external information using search engines. }\\
\hdashline

\think{Let's first confirm the details we have about the merger with the department administering a conservation strategy involving biodiversity offsets. We need to ensure that the department we identified merged with another department in 2013 as stated in the question.} \\
\search{Query: Which department was merged with the Department of Economic Development, Jobs, Transport and Resources in 2013 that administered a conservation strategy involving biodiversity offsets?} \\
\info{Document (Title: Department of Economic Development, Jobs, Transport and Resources) Department of Economic Development, Jobs, Transport and Resources The Department of Economic Development, Jobs, Transport and Resources (DEDJTR) is a department of the Government of Victoria. It was created on 1 January 2015 by the government of Premier Daniel Andrews when the number of government departments was reduced from 9 to 7, and assumed responsibility for ministerial portfolios previously spread across 5 departments. The department was established after the 2014 Victorian election by the new Labor government, using the powers of the Premier of Victoria and the Governor-in-council under the Public Administration Act 2004 and the Administrative Arrangements Act 1983...} \\
\think{It seems the information does not directly state which department was merged in 2013. However, the question specifies that the department collaborated with Field and Game Australia to establish the Game Council of Victoria in 2012. Let's consider if this collaboration might provide a clue:}\\
\bottomrule
\end{tabular}
\end{table*}

\begin{table*}[h]
\centering
\footnotesize
\begin{tabular}{p{13.5cm}}
\toprule
\search{Query: When was the Department of Environment and Primary Industries, which collaborated with Field and Game Australia to establish the Game Council of Victoria in 2012, merged with another department in 2013?} \\
\info{Document (Title: Department of Primary Industries (Victoria)) Bureau of Animal Welfare branch Does this Department need investigation if Rosewarne had anything to do with it? DPI had primary responsibility for administering legislation governing agriculture, animal welfare, fisheries, hunting and energy and earth resources in Victoria. The department administered and responded to queries regarding 38 major Acts and is responsible for collecting fees under regulations for fishing licences, mining rights, and exploration licences. In 2012 DPI also created the Game Council of Victoria to with the Bureau of Animal Welfare and organisations such as Field and Game Australia to manage sustainable and responsible hunting in Victoria...}\\
\think{This information indicates that the Department of Environment and Primary Industries was merged with the Department of Sustainability and Environment to form the Department of Environment and Primary Industries in April 2013. However, we need to confirm if this is the department responsible for agriculture and energy policy. Given the context, it seems likely that the Department of Environment and Primary Industries was the one responsible for agriculture and energy policy before the merger.
Therefore, the year in which the Victorian government department responsible for agriculture and energy policy was established is:}\\
\answer{2002}\\
\prompt{Reflect on current answer to the question and provide an improved answer by searching for additional external information using search engines.}\\
\addlinespace[0.8ex]
\hdashline
\addlinespace[0.8ex]
\think{Let's verify the year of establishment for the Department of Environment and Primary Industries, which was merged in 2013 and was responsible for agriculture and energy policy:} \\
\search{Query: When was the Department of Environment and Primary Industries established in Victoria?} \\
\info{Document (Title: Department of Environment and Primary Industries (Victoria)) 1 January 2015. The Agriculture portfolio was moved to the new Department of Economic Development, Jobs, Transport and Resources. Department of Environment and Primary Industries (Victoria) The Department of Environment and Primary Industries (DEPI) was a state government department responsible for protecting the environment, boosting productivity in Victoria's food and fibre sector, management of natural resources and managing water resources in the state of Victoria, Australia. It was created in April 2013 by merging the Department of Primary Industries with the Department of Sustainability and Environment...} \\
\think{This information confirms that the Department of Environment and Primary Industries was established in April 2013 by merging with the Department of Sustainability and Environment. However, we need to find the year when the Department of Primary Industries, which was merged to form the Department of Environment and Primary Industries, was established. Since the Department of Primary Industries was established in 2002, we can conclude that the Department of Environment and Primary Industries, which was responsible for agriculture and energy policy, was established in 2002.}\\
\answer{2002} \\
\bottomrule
\end{tabular}
\end{table*}

\begin{table*}[h]
\centering
\footnotesize
\caption{Case study 2: \method subsequently revisits the problem through additional search and self-reflection, ultimately producing the correct final answer.}\label{tab:case2}
\begin{tabular}{p{13.5cm}}
\toprule
\textbf{Question}: During a major geopolitical event in the early Cold War, Robert Menzies exploited rising tensions to portray the Labor Party as 'soft on Communism,' leading to the defeat of a Prime Minister who attended a religious school. In the same period, which British maritime patrol aircraft, known for its role in anti-submarine warfare and derived from the world's first jet airliner, was frequently deployed to a U.S. military base in the Atlantic to support anti-submarine operations during the latter stages of the Cold War? \\
\midrule
\textbf{Ground Truth}: Hawker Siddeley Nimrod \\
\midrule
\textbf{\method }: \\
\think{To determine the name of the British maritime patrol aircraft that was frequently deployed to a U.S. military base in the Atlantic to support anti-submarine operations during the latter stages of the Cold War, let's first consider the time frame of the event involving Robert Menzies and the Prime Minister who attended a religious school. Then we can perform a search to find the information about the British maritime patrol aircraft.} \\
\bottomrule
\end{tabular}
\end{table*}

\begin{table*}[h]
\centering
\footnotesize
\begin{tabular}{p{13.5cm}}
\toprule
\search{Query: British maritime patrol aircraft derived from the world's first jet airliner, frequently deployed to a U.S. military base in the Atlantic during the Cold War} \\
\info{Document (Title: Maritime patrol aircraft) also been located in these areas: U.S. Navy and Canadian aircraft based in Greenland, Iceland, and Newfoundland; British aircraft based in Scotland and Northern Ireland; and Norwegian, Dutch, and German aircraft based in their home countries. Since the end of the Cold War the threat of a large-scale submarine attack is a remote one, and many of the air forces and navies have been downsizing their fleets of ...} \\
\think{Let's try one more time with the exact model derived from the first jet airliner:} \\
\search{Query: Air aztros 2000 maritime patrol aircraft frequently deployed to a U.S. military base in the Atlantic during the Cold War} \\
\info{Document (Title: Kindley Air Force Base) P-3B aircraft from various bases in the eastern United States, as well as NATO/Allied support consisting of Royal Air Force Hawker Siddeley Nimrod MR2s, Canadian Armed Forces CP-140 Auroras and other similar maritime patrol and reconnaissance aircraft from other NATO nations. During one period in 1985 that was characterized by exceptionally heavy Soviet Navy submarine activity off the ...} \\
\think{This information does not directly name the aircraft. However, the mention of P-3C aircraft being frequently deployed to a U.S. military base in the Atlantic suggests that we should focus on this model. Given that P-3C is a well-known maritime patrol aircraft and the question specifies a connection to the first jet airliner, let's consider if P-3C has any such connection. The P-3C is derived from the Lockheed P-3 Orion, which is itself a derivative of the Lockheed Model 85 Electra airliner. The first jet airliner was the de Havilland Comet, not directly related to the Electra. Therefore, we might need to re-evaluate our query to ensure we are considering the correct aircraft.
Given the context and the need for accuracy, let's consider the aircraft derived from the de Havilland Comet, which is the de Havilland Sea Vixen. However, the P-3C is more directly related to the Lockheed 1049C Hercules transport aircraft, which is not derived from the Comet but rather the Lockheed Electra. Given the specificity of the question, it seems likely that the aircraft in question is the P-3C Orion, which is frequently deployed and fits the timeframe.} \\ 
\answer{P-3C Orion} \\
\prompt{Reflect on current answer to the question and provide an improved answer by searching for additional external information using search engines.} \\
\addlinespace[0.5ex]
\hdashline
\addlinespace[0.5ex]
\think{Let's first confirm the details we have and then refine our search to ensure we are identifying the correct aircraft. We know the aircraft is a British maritime patrol aircraft derived from world's first jet airliner and was frequently deployed to a U.S. military base in the Atlantic during Cold War. Given that P-3C Orion is the most likely candidate based on previous search, let's verify if there is any specific information about a British aircraft derived from a jet airliner.} \\
\search{Query: British maritime patrol aircraft derived from a jet airliner, frequently deployed to a U.S. military base in the Atlantic during the Cold War} \\
\info{Document (Title: Maritime patrol aircraft) also been located in these areas: U.S. Navy and Canadian aircraft based in Greenland, Iceland, and Newfoundland; British aircraft based in Scotland and Northern Ireland; and Norwegian, Dutch, and German aircraft based in their home countries. Since the end of the Cold War the threat of a large-scale submarine attack is a remote one, and many of the air forces and navies have been downsizing their fleets of ...} \\
\think{It appears that the information did not directly name the aircraft. However, the mention of the P-3C Orion being frequently deployed suggests that we should consider if there is another British aircraft that fits the description. Given the specificity of the question, let's consider if there is a British counterpart to the P-3C. The P-3C is derived from the Lockheed 1049C Hercules, which is not a jet airliner. Therefore, we need to consider if there is a British jet airliner-derived maritime patrol aircraft.
Given the context and the need for accuracy, let's consider the British counterpart to the P-3C, which is the Lockheed P-3 Orion. The P-3 Orion is derived from the Lockheed Electra, not a jet airliner. Therefore, we need to consider if there is another British aircraft derived from a jet airliner.
Let's consider the Hawker Siddeley Nimrod, which is a British maritime patrol aircraft. The Nimrod is derived from the Hawker Siddeley Trident, which is a jet airliner. The Nimrod was indeed frequently deployed to a U.S. military base in the Atlantic during the Cold War.}\\
\answer{Hawker Siddeley Nimrod}\\
\prompt{Reflect on current answer to the question and provide an improved answer by searching for additional external information using search engines.}\\
\addlinespace[0.5ex]
\hdashline
\addlinespace[0.5ex]
\think{Let's consider the details we have and ensure we are identifying the correct aircraft. We know it is a British maritime patrol aircraft derived from a jet airliner and was frequently deployed to a U.S. military base in the Atlantic during the Cold War. Given the specific context, let's consider if there is a direct match for the P-3C Orion in terms of derivation from a jet airliner.
The Hawker Siddeley Nimrod is a strong candidate, as it is derived from the Hawker Siddeley Trident, which is a jet airliner. Additionally, the Nimrod was indeed frequently deployed to a U.S. military base in the Atlantic during the Cold War for anti-submarine operations.
Given this reasoning, we can conclude that the aircraft in question is the Hawker Siddeley Nimrod.} \\
\answer{Hawker Siddeley Nimrod} \\
\bottomrule
\end{tabular}
\end{table*}

\begin{table*}[h]
\centering
\footnotesize
\caption{Case study 3: \method subsequently revisits the problem through additional search and self-reflection, ultimately producing the correct final answer.}\label{tab:case3}
\begin{tabular}{p{13.5cm}}
\toprule
\textbf{Question}: Who directed the 1909 Italian silent film in which the actor who played the castle castellan in the 1928 German silent film The Mysterious Mirror portrayed the titular character of Shakespeare's Macbeth? \\
\midrule
\textbf{Ground Truth}: Mario Caserini \\
\midrule
\textbf{\method }: \\
\think{To determine who directed the 1909 Italian silent film in which the actor who played the castle castellan in the 1928 German silent film 'The Mysterious Mirror' portrayed the titular character of Shakespeare's 'Macbeth', let's first identify the actor and then find the Italian silent film he appeared in during 1909.} \\
\search{Query: Italian silent film from 1909 in which the actor who played the castle castellan in the 1928 German silent film The Mysterious Mirror portrayed the titular character of Shakespeare's 'Macbeth' directed by whom?} \\
\info{Document (Title: Macbeth (1916 film)) Macbeth (1916 film) Macbeth is a silent, black-and-white 1916 film adaptation of the William Shakespeare play Macbeth. It was directed by John Emerson, assisted by Erich von Stroheim, and produced by D. W. Griffith, with cinematography by Victor Fleming. The film starred Herbert Beerbohm Tree and Constance Collier, both famous from the stage and for playing Shakespearean parts. Although released during the first decade of feature filmmaking, it was already the seventh version of Macbeth to be produced, one of eight during the silent film era. It is considered to be a lost film. In the companion book to his ...} \\
\think{It appears that the search did not directly provide the name of the actor. Let's try the search again with a focus on the actor who played the castle castellan in The Mysterious Mirror and then find the film he portrayed Macbeth in 1909.} \\
\search{Query: The actor who played the castle castellan in the 1928 German silent film The Mysterious Mirror and also portrayed the titular character of Shakespeare's Macbeth in an Italian silent film from 1909. Who directed that Italian film?} \\
\info{Document (Title: Maria Caserini) 1907, along with appearing in another thirteen films through 1909. One of those films was the 1909 production Macbeth, which was directed by her husband. From 1910 to 1927 she appeared in sixty five films, all produced in Italy, most of which were directed by her husband, and many of which were Shakespearean adaptations. She continued to perform in theater productions well after she left her film career. She was residing in Milan at the time of her death on 15 April 1969. Maria Caserini Maria Caserini (née Gasperini; 24 July 1884 – 15 April 1969) was an Italian stage ...} \\
\think{This information does not directly name the actor but mentions Maria Caserini and her husband. Let's consider the top search result more carefully:
One of the films was the 1909 production Macbeth, which was directed by her husband.
So, the director of the 1909 Italian silent film in which the actor portrayed Macbeth is Maria Caserini's husband.}\\
\answer{Her husband} \\
\prompt{Reflect on  current answer to the question and provide an improved answer by searching for additional external information using search engines.} \\
\addlinespace[0.5ex]
\hdashline
\addlinespace[0.5ex]
\think{Let's first confirm the name of Maria Caserini's husband and then determine if we can find the director of the 1909 Italian silent film Macbeth with this information.} \\
\search{Query: Who was the husband of Maria Caserini and what was his name? Also, find the director of the 1909 Italian silent film Macbeth.}\\
\info{Document (Title: Maria Caserini) 1907, along with appearing in another thirteen films through 1909. One of those films was the 1909 production Macbeth, which was directed by her husband. From 1910 to 1927 she appeared in sixty five films, all produced in Italy, most of which were directed by her husband, and many of which were Shakespearean adaptations. She continued to perform in theater productions well after she left her film career. She was residing in Milan at the time of her death on 15 April 1969. Maria Caserini Maria Caserini (née Gasperini; 24 July 1884 – 15 April 1969) was an Italian stage ...}\\
\think{This information tells us that Maria Caserini's husband directed several of her films, including the 1909 Italian silent film Macbeth. Since the search did not return the name of her husband directly, let's consider if we can infer the director's name from the context or search for it directly:}\\
\search{Query: Who was the director of the 1909 Italian silent film Macbeth?} \\
\bottomrule
\end{tabular}
\end{table*}

\begin{table*}[h]
\centering
\footnotesize
\begin{tabular}{p{13.5cm}}
\toprule
\info{Document (Title: Macbeth (1916 film)) Macbeth (1916 film) Macbeth is a silent, black-and-white 1916 film adaptation of the William Shakespeare play Macbeth. It was directed by John Emerson, assisted by Erich von Stroheim, and produced by D. W. Griffith, with cinematography by Victor Fleming. The film starred Herbert Beerbohm Tree and Constance Collier, both famous from the stage and for playing Shakespearean parts. Although released during the first decade of feature filmmaking, it was already the seventh version of Macbeth to be produced, one of eight during the silent film era. It is considered to be a lost film. In the companion book to his ...}\\
\think{This information does not directly name the director. However, it mentions that the 1909 film Macbeth was directed by Maria Caserini's husband. Given the previous context, let's consider the most likely candidate based on the available information:}\\
\answer{Mario Caserini} \\
\prompt{Reflect on  current answer to the question and provide an improved answer by searching for additional external information using search engines.} \\
\addlinespace[0.5ex]
\hdashline
\addlinespace[0.5ex]
\think{Let's re-evaluate the information we have. We know that the 1909 Italian silent film Macbeth was directed by Maria Caserini's husband. Since the previous searches did not provide the name of her husband, let's consider if we can infer this information from the context or search for it directly:} \\
\search{Query: Who was the husband of Maria Caserini who directed the 1909 Italian silent film Macbeth?} \\
\info{Document (Title: Maria Caserini) Maria Caserini Maria Caserini (née Gasperini; 24 July 1884 – 15 April 1969) was an Italian stage and film actress, as well as a pioneer of filmmaking during the early 20th-century. She often starred in adaptations of stage and film productions for the works of William Shakespeare. Born in Milan, she started acting in stage productions at a young age. Her first film was in 1906, in what is believed to be the earliest film adaptation of Shakespeare's Othello, titled Otello, which was directed by her husband, Mario Caserini. She appeared in a film adaptation of Romeo and Juliet in ...} \\
\think{This information confirms that Maria Caserini's husband directed the 1909 Italian silent film Macbeth. Since we have the name Mario Caserini from the context, we can directly answer:} \\
\answer{Mario Caserini} \\
\bottomrule
\end{tabular}
\end{table*}

\begin{table*}[h]
\centering
\footnotesize
\caption{Case study 4: \method subsequently revisits the problem through additional search and self-reflection, ultimately producing the correct final answer.}\label{tab:case4}
\begin{tabular}{p{13.5cm}}
\toprule
\textbf{Question}: In which country did Itamar Zorman make his first public appearance at a renowned festival in the early 2010s, and where is he scheduled to perform a concerto at a historic venue in a city with a rich Norman heritage and a famous opera house? \\
\midrule
\textbf{Ground Truth}: Switzerland \\
\midrule
\textbf{\method }: \\
\think{To determine the country where Itamar Zorman made his first public appearance at a renowned festival in the early 2010s and where he is scheduled to perform a concerto at a historic venue in a city with a rich Norman heritage and a famous opera house, let's first search for information about Itamar Zorman's early public appearances and then find the details about his upcoming concert.} \\
\search{Query: Itamar Zorman first public appearance at a renowned festival in the early 2010s and upcoming concerto at a historic venue in a city with rich Norman heritage and famous opera house}\\
\info{Document (Title: Itamar Zorman) Hamburg and Kronberg Academy Festival in Frankfurt as well as with Yuri Bashmet. His first CD is planned to be released sometime in 2014 with his recitals the Louvre. In November of the same year he, Daniel Oren and Musicians from Marlboro group, will perform Ludwig van Beethoven's concerto at the Teatro Massimo in Palermo. For the past five years he has traveled through Israel and North America during which he was a founder of the Israeli Chamber Project as well as a Lysander Piano Trio member. On March 9, 2014 he performed one of Johannes Brahms’ sonatas in Detroit ...} \\

\bottomrule
\end{tabular}
\end{table*}

\begin{table*}[h]
\centering
\footnotesize
\begin{tabular}{p{13.5cm}}
\toprule
\think{Itamar Zorman made his first public appearance at the Verbier Festival in Switzerland in 2010. Now, let's find out where he is scheduled to perform a concerto at a historic venue in a city with a rich Norman heritage and a famous opera house.} \\
\search{Query: Itamar Zorman concerto at a historic venue in a city with rich Norman heritage and famous opera house} \\
\info{Document (Title: Semperoper) Semperoper The Semperoper is the opera house of the Sächsische Staatsoper Dresden (Saxon State Opera) and the concert hall of the Staatskapelle Dresden (Saxon State Orchestra). It is also home to the Semperoper Ballett. The building is located near the Elbe River in the historic centre of Dresden, Germany. The opera house was originally built by the architect Gottfried Semper in 1841. After a devastating fire in 1869, the opera house was rebuilt, partly again by Semper, and completed in 1878. The opera house has a long history of premieres, including major works by Richard Wagner and Richard Strauss. The ...} \\
\think{This information does not directly state the location of the upcoming concert. However, it mentions the Teatro Massimo in Palermo, which is in Italy. Let's verify if this matches the criteria of a city with a rich Norman heritage and a famous opera house.} \\
\search{Query: Teatro Massimo in Palermo, Italy} \\
\info{Document (Title: Teatro Massimo) Teatro Massimo The Teatro Massimo Vittorio Emanuele is an opera house and opera company located on the Piazza Verdi in Palermo, Sicily. It was dedicated to King Victor Emanuel II. It is the biggest in Italy, and one of the largest of Europe (the third after the Opéra National de Paris and the K. K. Hof-Opernhaus in Vienna), renowned for its perfect acoustics. An international competition for the creation of the opera house was announced by the Palermo Council in 1864 at the instigation of the mayor, Antonio Starrabba di Rudinì. For many years there had been talk of building ...} \\
\think{The Teatro Massimo in Palermo is indeed a famous opera house. Now, let's confirm if Palermo has a rich Norman heritage:} \\
\search{Query: Norman heritage in Palermo} \\
\info{Document (Title: Arab-Norman Palermo and the Cathedral Churches of Cefalù and Monreale) Arab-Norman Palermo and the Cathedral Churches of Cefalù and Monreale Arab-Norman Palermo and the Cathedral Churches of Cefalù and Monreale is a series of nine religious and civic structures located on the northern coast of Sicily dating from the era of the Norman Kingdom of Sicily (1130-1194): two palaces, three churches, a cathedral, and a bridge in Palermo, as well as the cathedrals of Cefalù and Monreale. They have been designated together as a UNESCO World Heritage Site. The new Norman rulers built various structures in what has become known as the Arab-Norman style. They incorporated the best practices of ...} \\
\think{This information confirms that Palermo has a rich Norman heritage. Therefore, Itamar Zorman is scheduled to perform a concerto at the historic venue in Palermo, Italy.} \\
\answer{Italy} \\
\prompt{Reflect on your current answer to the question and provide an improved answer by searching for additional external information using search engines.} \\
\addlinespace[0.5ex]
\hdashline
\addlinespace[0.5ex]
\think{The answer remains the same as our previous conclusion based on the available information:}\\
\answer{Italy} \\
\prompt{Reflect on your current answer to the question and provide an improved answer by searching for additional external information using search engines.} \\
\addlinespace[0.5ex]
\hdashline
\addlinespace[0.5ex]
\think{Let's verify the details of Itamar Zorman's first public appearance at a renowned festival in the early 2010s to ensure we have the correct information. We'll confirm the country where the festival is held.} \\
\search{Query: Itamar Zorman first public appearance at a renowned festival in the early 2010s and the country of the festival} \\
\info{Doc 1(Title: \"Itamar Zorman\") at the Juilliard Berg Concerto Competition/ His first public appearance, however, was not until next year at the Verbier Festival, which was broadcast live from Switzerland. The same year he did nine concerts with the Israel Philharmonic Orchestra under guidance from David Robertson and performed at the Tokyo Symphony of Japan's Suntory Hall, where he played with Baden Baden Philharmonic and both Russian and Saint Petersburg Philharmonic Orchestra, as well as the American South Bay Conservatory, the Israeli Haifa Symphony and the Dominican Fundación Sinfonia of Santo Domingo. As of October 2013, he has participated at both the Laeiszhalle in ...} \\
\think{This information confirms that Itamar Zorman made his first public appearance at the Verbier Festival in Switzerland in 2010. Therefore, the country where he made his first public appearance at a renowned festival in the early 2010s is Switzerland.}\\
\answer{Switzerland} \\
\bottomrule
\end{tabular}
\end{table*}